# Sentiment Expression via Emoticons on Social Media

Hao Wang, Jorge A. Castanon
Silicon Valley Lab
IBM
San Jose, USA
{haowang, jorgecasta}@us.ibm.com

*Abstract*—Emoticons (e.g., :) and :( ) have been widely used in sentiment analysis and other NLP tasks as features to machine learning algorithms or as entries of sentiment lexicons. In this paper, we argue that while emoticons are strong and common signals of sentiment expression on social media the relationship between emoticons and sentiment polarity are not always clear. Thus, any algorithm that deals with sentiment polarity should take emoticons into account but extreme caution should be exercised in which emoticons to depend on. First, to demonstrate the prevalence of emoticons on social media, we analyzed the frequency of emoticons in a large recent Twitter data set. Then we carried out four analyses to examine the relationship between emoticons and sentiment polarity as well as the contexts in which emoticons are used. The first analysis surveyed a group of participants for their perceived sentiment polarity of the most frequent emoticons. The second analysis examined clustering of words and emoticons to better understand the meaning conveyed by the emoticons. The third analysis compared the sentiment polarity of microblog posts before and after emoticons were removed from the text. The last analysis tested the hypothesis that removing emoticons from text hurts sentiment classification by training two machine learning models with and without emoticons in the text respectively. The results confirms the arguments that: 1) a few emoticons are strong and reliable signals of sentiment polarity and one should take advantage of them in any sentiment analysis; 2) a large group of the emoticons conveys complicated sentiment hence they should be treated with extreme caution.

*Keywords-emoticon; sentiment; polarity; Twitter; social media*

I. INTRODUCTION

Emoticons, such as :) ;) :-) and :(, are frequently used online in social media, IM (e.g., Skype), blogs, forums, and other kinds of online social interactions. Because they are commonly used in online communications and they are often direct signals of sentiment, emoticons in text were widely used by NLP researchers in tasks such as sentiment analysis as features to machine learning algorithms or as entries of sentiment lexicons for rule-based approaches.

Different online communities and tools may elicit varied degrees of emoticon usage. Twitter, a microblogging site, is one of most popular social media. For researchers and businesses, having access to its huge amount of user-generated data is critical for understanding user behavior and the sentiment expressed. With access to about 50 million tweets per day (through the Twitter Decahose API), we thought it would be interesting to understand the prevalence of emoticons on Twitter nowadays, how users express and perceive sentiment through emoticons, and whether emoticons can be used as a reliable cue for sentiment polarity classification.

II. PREVIOUS WORK

There has been abundant studies on sentiment analysis in recent years[1]. In particular, its application on social media posts has gathered a lot of interests in both academia and the industry [2-6].

In many of the past studies, emoticons played an important role in both building sentiment lexicons and in training machine learning classifiers [2, 7-12]. It has been thought that emoticons are reliable indicators of sentiment. Several attempts have been made to build sentiment corpus based on emoticons [10, 13, 14]. However, none of the studies has directly examined the relationship between emoticons and sentiment polarity on social media as well as the roles of emoticons in such context. This work aims to fill the gap and answer the following questions: 1) how prevalent are emoticons on Twitter today? 2) how are emoticons used and in what context? 3) what meaning do emoticons convey? 4) how do emoticons help in expressing sentiment? 5) are emoticons reliable cues for sentiment?

III. EMOTICONS IN SOCIAL MEDIA

Users of social media and IM tools use a variety of emoticons. Some of the emoticons, such as :) and :(, are widely used and many others are only used by a fraction of users. We compiled a relatively comprehensive list of 164 emoticons from previous studies and the Wikipedia [15, 16].

We then searched for any emoticon in that list in a very large Twitter data set that contains all the tweets collected through the Twitter Decahose API in the entire month of March 2015 (the Twitter Decahose API provides 10% of entire Twitter traffic). A tweet, which is one microblog post, contains 140 characters maximum. The data set contains roughly 1.5 billion tweets. 8,625,753 of emoticons were found in that data set. Majority of the tweets contain only one emoticon.

Table 1 lists the most frequent emoticons and their frequency in our data set. :) alone were used more than 43% of the times. As expected, many of the emoticons were used infrequently. For the rest of the analyses in this paper, we selected and used the emoticons that occurred more than 0.1%, which results in the 34 emoticons in Table 1. In summary, we have shown that emoticons, especially the few widely used ones, are prevalent in tweets.

TABLE 1. MOST FREQUENT EMOTICONS

| Icon | Frequency | % of Total | Icon | Frequency | % of Total |
|---|---|---|---|---|---|
| `:)` | 3760375 | 43.6% | `=)` | 54022 | 0.6% |
| `:(` | 1024100 | 11.9% | `;P` | 37909 | 0.4% |
| `;)` | 660071 | 7.7% | `:-D` | 34087 | 0.4% |
| `:D` | 628811 | 7.3% | `D:` | 31697 | 0.4% |
| `:-)` | 488580 | 5.7% | `;(` | 24706 | 0.3% |
| `:/` | 320630 | 3.7% | `8)` | 21114 | 0.2% |
| `(:` | 169264 | 2.0% | `:-/` | 19222 | 0.2% |
| `:P` | 159630 | 1.9% | `:|` | 17549 | 0.2% |
| `xD` | 155258 | 1.8% | `XP` | 15864 | 0.2% |
| `:')` | 145779 | 1.7% | `=D` | 15313 | 0.2% |
| `XD` | 136973 | 1.6% | `:]` | 13621 | 0.2% |
| `;-)` | 113814 | 1.3% | `D8` | 13322 | 0.2% |
| `:p` | 106509 | 1.2% | `DX` | 12893 | 0.1% |
| `:'(` | 95270 | 1.1% | `:-P` | 11324 | 0.1% |
| `:@` | 85173 | 1.0% | `=]` | 9542 | 0.1% |
| `:-(` | 79821 | 0.9% | `>:(` | 8849 | 0.1% |
| `;D` | 56948 | 0.7% | `:\` | 8486 | 0.1% |

## IV. EMOTICONS AND SENTIMENT

In sentiment analysis, polarity of sentiment (e.g., positive, negative or neutral) is of particular interest to researchers and business applications. However, the emotions expressed by the emoticons often cannot be captured by the three polarity categories. Many emoticons do not belong to exactly one of the categories. For example, `:/` is often used to express an emotional state of annoyed and uneasy, which could be an indication of negative sentiment for some people but neutral for others. To validate and quantify this intuition, we did a survey of emotions expressed by the 34 emoticons in Table 1. 31 participants completed the survey. Each participant was asked to choose one from the following four options for each emoticon: *Positive*, *Negative*, *Neutral*, and *None of the above or not sure*. Fig. 1 summarizes the results as the proportion of all responses for each option and emoticon. The results are very interesting while largely expected. There are two groups of emoticons that were labeled by majority of the participants as *positive* or *negative* (in the top and bottom of Fig. 1). Almost all participants agreed with little uncertainty that `:D` and `:)` are *positive* and `:-(`, `:'(` and `:(` are *negative*. However, a large number of emoticons was labeled with a mix of the three polarities as well as uncertain (in the middle of Fig. 1). For example, *positive* has the most responses for `:')` but *negative* and *None of the above or not sure* also have large shares of the responses, which reflects the complex and ambiguous nature of human emotion and language. It is also worth noting that even for some of the most common emoticons (e.g., `:-)` and `:p`) the interpretation of the emotions expressed is not perfectly consistent among the participants.

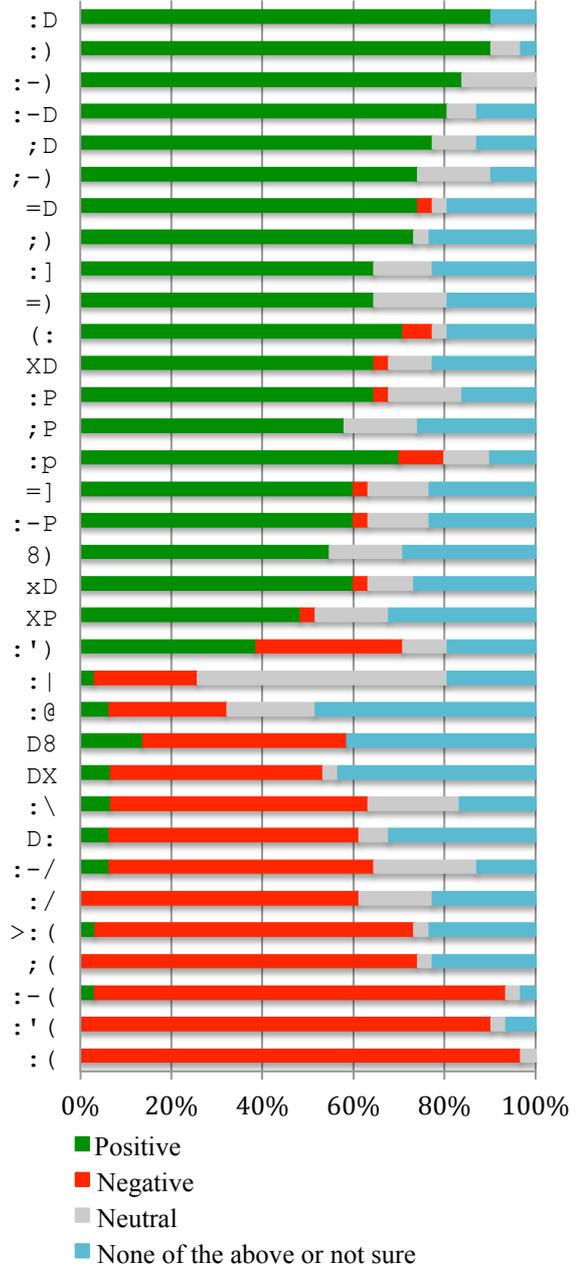

Figure 1. Survey of emotion expressed by emoticons

This survey of sentiment expressed by emoticons confirmed the hypothesis that some emoticons are reliable indicators of sentiment polarity while there is also large variation in how people express emotions through emoticons and how they interpret the sentiment conveyed by emoticons. Therefore, extreme caution should be exercised when utilizing the rich information in emoticons for better sentiment analysis. To further investigate the relationship between emoticons and sentiment, we carried out three more analyses to understand the contexts in which emoticons are used and the influence of emoticons on sentiment classification.

In the next three analyses, we drilled into a subset of the Twitter data: one day of tweets from March 4, 2015 where 276,207 emoticons were found from 45,169,774 tweets.

## A. Clustering emoticons and words

Interesting questions regarding the usage of emoticons are in what context they are usually used and what meaning emoticons convey. In this analysis, we applied two machine learning algorithms to answer these questions. Firstly, we used a version of word2vec [17, 18], an algorithm based on deep neural networks, to define the representation of the words, including emoticons in the data set. Secondly, we use the k-means algorithm to cluster the words so we can understand the exact meaning of the emoticons through the words that appear in the same cluster. Both of these machine learning algorithms help us define, explain and interpret emoticons in the context of Twitter data. Next, we describe the details about the word2vec and k-means algorithms, respectively.

For this experiment, all the tokens with a frequency less than 50 were filtered out. The number of features that word2vec generates was determined based on the decay of the singular values of the resulted feature matrix. Since small singular values explain little variation of the data, we chose a threshold of approximately 0.001 for the singular values to set the number of features to 500 (Fig. 2). The feature matrix generated by word2vec was of the size 4480 words by 500 features. For the k-means algorithm, we used k=50 following the rule in [19], in which the authors argue that k approx sqrt(n/2) is a reasonable choice of k where n is the number of observations.

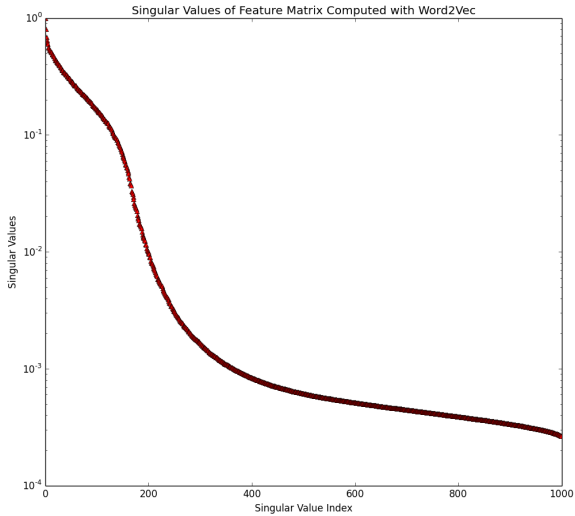

Figure 2. Singular values of word2vec feature matrix

Table 2 shows the emoticons in each cluster. Notice that clusters A, B, D and E include mostly positive emoticons and cluster C includes emoticons in the negative group. An interesting observation is that the emoticon `:|`, labeled as neutral by more than a half of the participants of the survey (Fig. 1), appears in the negative cluster C. This confirms that `:|` conveys a negative sentiment to some degree. Table 3 shows some sample words in each of the clusters with emoticons. The words help us understand the emoticons in the same clusters in terms of the sentiment they express. For example, `:')` received almost similar numbers of positive and negative responses in the survey. However, the words in cluster C suggests that `:')` appears to be used primarily in negative contexts. In summary, this analysis validates the idea that the emoticons were often used in consistent contexts to help express sentiment and the sentiment expressed by emoticons agrees well with words around them. Detailed analysis of the content of the clusters would be of particular interest to socio-linguistics researchers.

TABLE 2. CLUSTERS OF EMOTICONS

| A | `:) :D =)` |
|---|---|
| B | `;) :-) ;-) :-D =D ;P =]` |
| C | `:( :/ XD :') :'( :-( D: ;( :-/ :| :\` |
| D | `:P ;D :-P :] :p` |
| E | `(:` |
| F | `XP` |
| G | `DX` |
| H | `8)` |

TABLE 3. SAMPLE WORDS FROM CLUSTERS WITH EMOTICONS

| A | good thanks happy birthday fantastic lovely wonderful amazing beautiful welcome awesome congrats |
|---|---|
| B | smile friends face music favorite pic heart kind coffee sexy pleasure positive exciting healthy |
| C | miss sorry bad hate sad omg shit fuck sick late bitch mad ugh ugly broke |
| D | what don't no know think can't lol why ever never look feel |
| E | love follow please hey wish goodnight dream iloveyou |
| F | stuck shoot fatally |
| G | #music camera smartphone taylor swift |
| H | best fun coming week year playing top happiness friday spring weekend summer movie party happiest |

## B. Sentiment expression with vs. without emoticons

To understand the degree of which emoticons were used to express sentiment, we carried out this analysis to compare the sentiment of a tweet before and after emoticons are removed from the text. Such a comparison would reveal how much emoticons were relied on to express sentiment in a tweet.

500 tweets were randomly selected from the tweets with emoticons from March 4, 2015. They were manually annotated with one of the three labels: *positive sentiment*, *negative sentiment* or *other*. A tweet was labeled as *positive* or *negative* if the sentiment is clear from reading the text (including the emoticons). It was labeled as *other* if it does not express any sentiment (*neutral*), expresses multiple sentiments, or the sentiment is not clear. The same set of tweets was manually annotated again using the three labels after the emoticons were removed from the text.

By comparing the changes in the labels, we can understand how exactly emoticons help express sentiment in tweets. Table 4 shows the confusion matrix of the labels. In the original tweets, 343 and 101 out of 500 were identified as *positive* and *negative*, respectively. 56 are labeled as *Other*. After removing the emoticons in the tweets, only 165 and 57 out of 500 were identified as *positive* and *negative*, respectively. Those are nearly half of the origi-nal numbers. More than half of the tweets (278) fell into the *other* category. This result suggests that in about a half of the cases emoticons were the only signal of sentiment and in the other half of the cases emoticons were facilitating the expression of sentiment. Therefore, emoticons are a crucial component (sometimes the only component) of sentiment expression in short microblog posts. It is extremely important for any sentiment analysis of such kind of text to take emoticons into account. Otherwise, the recall of the sentiment analysis may suffer because of missing a significant portion of the documents.

TABLE 4. CONFUSION MATRIX OF SENTIMENT LABELS WITH VS. WITHOUT EMOTICONS

|  |  | Without Emoticons | | | |
| --- | --- | --- | --- | --- | --- |
|  |  | Negative | Positive | Other | Total |
| With Emoticons | Negative | 54 | 3 | 44 | 101 |
|  | Positive | 3 | 162 | 178 | 343 |
|  | Other | 0 | 0 | 56 | 56 |
|  | Total | 57 | 165 | 278 | 500 |

*C. Classifying sentiment with vs. without emoticons*

The results from last analysis provided support for the importance of emoticons in sentiment expression. One may wonder if removing emoticons from text will actually hurt sentiment classification using machine learning algorithms. To test this hypothesis, we trained two Naive Bayes classifiers using the bag-of-words model. The manually annotated 500 tweets in subsection B were used as the training and test data. One classifier was trained and tested on the original tweets. Another classifiers was trained and tested on the same set of tweets with emoticons removed. We used 5 fold cross-validation for the training and testing of both classifiers. Standard metrics including precision, recall, F1 and accuracy were used to evaluate the classifiers.

Table 5 shows the metrics averaged over the five runs. Classifiers trained with the original tweets, which include the emoticons, are reasonably accurate giving the training data was only 400 samples. Classifiers trained on tweets with emoticons removed have much lower accuracy, precision and recall for the *positive* and *negative* classes. For the *other* class, the models performed better in the without-emoticon condition presumably because the *other* class now contains more training samples.

There could be two potential explanations for the differences in precision and recall for *positive* and *negative* classes. Firstly, it could be that removing the emoticons was hurting the classifiers because less information was available to it. Secondly, it may be due to the fact that the *positive* and *negative* classes had less training samples in the without-emoticon condition. Although this experiment could be improved by annotating more samples, current results seem to support that argument that removing emoticons from text hurts the performance of machine learning classifiers as well.

V. CONCLUSIONS AND FUTURE WORK

In this paper, we have shown that emoticons are widely used by Twitter users. In particular, emoticons expressing positive sentiment, such as `:)` and `;)`, were the dominant majority on Twitter. A group of emoticons expressing negative sentiment was commonly used too, while many others were used relatively infrequently. This observation is indeed in line with the Zipf's law of word frequencies [20]. We conducted a survey to understand the perception of sentiment polarity of the emoticons by human, which revealed that some emoticons are strong and reliable signals of sentiment polarity while many others inherited the complexity and ambiguity in human language and emotion. We carried out three more analyses using one of day of the Twitter data and investigated the relationship between emoticons and sentiment expression on the popular social media. The first analysis illustrated the fact that emoticons are used consistently in similar contexts by demonstrating that emoticons and words expressing similar sentiment are grouped into same clusters. It also showed that the complex meaning conveyed by some of the emoticons, such as `:|` and `:')`, can be understood through the words that appeared in the same contexts. The second analysis, comparing sentiment of tweets with and without emoticons, provided direct evidence on the importance of emoticons in expressing sentiment on social media. In nearly a half of the cases, emoticons were the only component in the text that expressed some positive or negative sentiment. When the emoticons were removed, the sentiment of those tweets became neutral or unclear. In the last analysis, we assessed the impact of removing emoticons from text to machine learning classifiers. We showed that the classifiers became less accurate when emoticons were removed with the caveat that the classifier was trained with less positive and negative samples. The results from the analyses together confirms the arguments that: 1) a few emoticons are strong and reliable (and sometimes unique) signals of sentiment polarity and one should take advantage of them in any sentiment analysis; 2) a large group of the emoticons conveys complicated sentiment hence they should be treated with extreme caution. In conclusion, this study directly examined the relationship between emoticons and sentiment polarity and provides important recommendations for developing future sentiment analysis algorithms and solutions. Some of the methods and results in this study may also be informative for socio-linguistics researchers interested in emoticon usage and sentiment expression on social media.

TABLE 5. EVALUATION OF THE CLASSIFIERS

|  |  | Positive | | | Negative | | | Other | | |
| --- | --- | --- | --- | --- | --- | --- | --- | --- | --- | --- |
|  | Accuracy | Precision | Recall | F1 | Precision | Recall | F1 | Precision | Recall | F1 |
| With Emoticon | 0.78 | 0.84 | 0.87 | 0.86 | 0.68 | 0.69 | 0.68 | 0.47 | 0.38 | 0.40 |
| Without Emoticon | 0.61 | 0.54 | 0.56 | 0.54 | 0.56 | 0.47 | 0.51 | 0.68 | 0.68 | 0.68 |

For future work, we plan to annotate a larger set of tweets to corroborate the results obtained in Section IV.B and IV.C.